\def\year{2021}\relax
\documentclass[letterpaper]{article} 
\usepackage{aaai21}  
\usepackage{times}  
\usepackage{helvet} 
\usepackage{courier}  
\usepackage[hyphens]{url}  
\usepackage{graphicx} 
\usepackage{natbib}  
\usepackage{caption} 
\usepackage{epsfig}
\usepackage[colorlinks]{hyperref}
\usepackage{graphicx}
\usepackage{amsmath}
\usepackage{amssymb}
\usepackage{soul}
\usepackage{url}
\usepackage{graphicx}
\usepackage{float}
\usepackage{amsthm}
\usepackage{booktabs}
\usepackage{makecell}
\usepackage[switch]{lineno}
\usepackage{algorithmic}
\usepackage{subfig}
\usepackage{bm}
\usepackage[ruled,vlined,linesnumbered,noresetcount]{algorithm2e}

\newcommand{\img}{\textbf{I}}

\newcommand{\etal}{\textit{et al.}}

\newcommand{\la}{y}

\newcommand{\net}{f}
\newcommand{\pool}{S}

\newcommand{\eg}{\emph{e.g.}}

\def\ie{\emph{i.e.}}

\title{Beating Attackers At Their Own Games:\\ Adversarial Example Detection Using Adversarial Gradient Directions}
\author{Yuhang Wu, Sunpreet S. Arora, Yanhong Wu, and Hao Yang\\}
\affiliations{
Visa Research\\
\{yuhawu, sunarora, yanwu, haoyang\}@visa.com\\
}

\begin{document}

\maketitle

\begin{abstract}
Adversarial examples are input examples that are specifically crafted to deceive machine learning classifiers. State-of-the-art adversarial example detection methods characterize an input example as adversarial either by quantifying the magnitude of feature variations under multiple perturbations or by measuring its distance from estimated benign example distribution. Instead of using such metrics, the proposed method is based on the observation that the directions of adversarial gradients when crafting (new) adversarial examples play a key role in characterizing the adversarial space. Compared to detection methods that use multiple perturbations, the proposed method is efficient as it only applies a single random perturbation on the input example. Experiments conducted on two different databases, CIFAR-10 and ImageNet, show that the proposed detection method achieves, respectively, 97.9\% and 98.6\% AUC-ROC (on average) on five different adversarial attacks, and outperforms multiple state-of-the-art detection methods. Results demonstrate the effectiveness of using adversarial gradient directions for adversarial example detection.
\end{abstract}

\section{Introduction}
Deep neural networks (DNNs) are being widely used in classification systems because of their exceptional performance on a wide range of practical problems from fraud detection to biometrics. However, recent research in adversarial machine learning has highlighted a major security concern with use of the DNNs in practical applications. Researchers have shown that an adversary can add human-imperceptible malicious perturbations to input examples to cause incorrect predictions \cite{DBLP:journals/corr/SzegedyZSBEGF13,DBLP:conf/sp/Carlini017}. Such input examples are termed \textit{adversarial examples}.

\begin{figure}[!t]
	\centering
	\includegraphics[width=7.4cm, height=3.3cm]{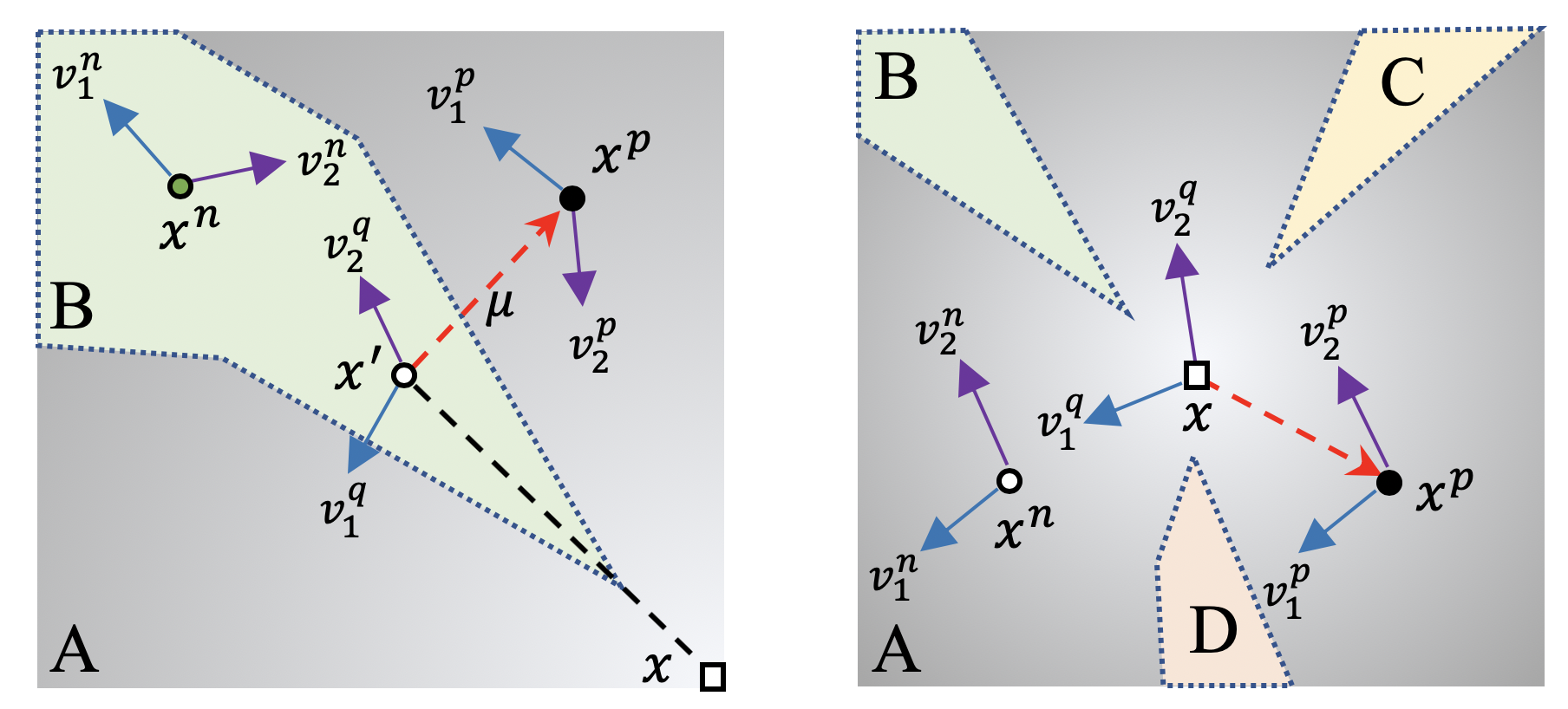}
	\caption{Schematic illustration of the proposed approach in the 2D embedding space (L) of  adversarial example $x'$, and (R) of benign example $x$. Assume that an attacker crafts an adversarial example $x'$ from a benign example $x$ of class A, and aims to deceive a deep network into misclassifying $x'$ in class B. Given an input example $x^{q}$, the proposed method perturbs $x^{q}$ using a random perturbation $\mu$ to obtain $x^{p}$, and retrieves $x^{q}$'s nearest neighbor prototype $x^n$ from a reference database, which shares the same predicted class with $x^{q}$. Following this, adversarial gradients of $x^{q}, x^{p}$, and $x^{n}$ (denoted as $v^i_k$) corresponding to their top $K$ predicted classes (here $K$=2) are calculated. Pairwise angular similarity between $\{v^q_{k}, v^p_{k}\}, \{v^q_{k}, v^n_{k}\}$, and $\{v^p_k, v^n_k\}$ is computed ($k=\{1,2\}$) and used to determine if $x^{q}$ is adversarial. Note that adversarial gradient directions in (R) are comparatively consistent and have larger angular similarities than (L).}
	\label{fig:1}
\end{figure}

State-of-the-art adversarial example detection methods primarily exploit the following two observed properties of adversarial examples: (i) adversarial examples are comparatively more sensitive to perturbations in the input space than benign examples, and (ii) the distance of an adversarial example to the data distribution of benign examples is anomalous.

The methods presented in \cite{DBLP:journals/corr/abs-1805-05010, DBLP:conf/ndss/Xu0Q18, DBLP:conf/aaai/TianYC18} are based on observation (i). These methods transform an input example using geometrical transformations (\eg, rotation, translation) and/or filtering operations (\eg median filtering). The changes in prediction probabilities of the classifier before and after transformation are aggregated and compared to determine if the input example is adversarial. Despite promising results, one limitation of these methods is the use of multiple random transformations which, in turn, increases the computational complexity. The proposed method, on the other hand, applies a single random perturbation to an input example to determine if the example is adversarial by using adversarial gradient directions. 

Instead of relying on observation (i), the method described in \cite{DBLP:journals/corr/FeinmanCSG17} is based on observation (ii) and uses kernel density estimation to identify low probability adversarial sub-spaces. The method reported in \cite{DBLP:conf/iclr/Ma0WEWSSHB18} characterizes the space-filling capability of the region surrounding an adversarial example using local intrinsic dimensions, and the one presented in \cite{DBLP:conf/nips/LeeLLS18} models the likelihood of an input example being adversarial (out-of-distribution) using class-conditional Gaussian distributions. A key assumption in these approaches is the availability of sufficient predicted class examples in the reference database to estimate the distribution of benign examples. In contrast, the proposed method uses a single benign example from the predicted class to determine whether an input example is adversarial. This is quite useful in the few shot learning setting, especially when the number of predicted class examples in the reference database is limited.

The proposed method is inspired by the research presented in \cite{DBLP:conf/icml/RothKH19}, which shows that for an adversarial input, adding random perturbations to the input space induces directional feature variations. On the contrary, the feature variations induced by random perturbations on a benign input are non-directional. Roth \etal \cite{DBLP:conf/icml/RothKH19} modeled the statistics of directional feature variations by measuring the changes in logits between each pair of predicted class examples for different random perturbations. In contrast, our approach characterizes feature variations in a more intuitive manner. Our findings show that the gradient directions used to craft ``adversarial examples" are sufficiently discriminative to characterize the adversarial input space. We observe that benign examples usually have similar adversarial gradient directions (AGDs) before and after a random perturbation (RP), while the difference in AGDs of adversarial examples before and after a RP is significant. We also observe that AGDs strongly depend on the original class of adversarial examples, and that the examples that are closer on the data manifold share similar AGDs. Given these observations, we compute AGDs of the nearest predicted class example to the input example (prototype). The AGDs of the input (query) example are compared with the AGDs of the prototype as well as the randomly perturbed example as depicted in \mbox{Fig. \ref{fig:1}}. Finally, a classifier is trained using the angular similarities between each pair of the computed AGDs to detect adversarial examples.

The main contributions of this work are as follows:
\begin{itemize}
    \item Use of Adversarial Gradient Directions (AGDs) for adversarial example detection. To the best of our knowledge, this is the first work that uses AGDs for adversarial example detection. AGDs are typically used for crafting adversarial examples.
    \item Demonstration of adversarial example detection using a single random perturbation and a single example from the predicted class.
    \item State-of-the-art adversarial example detection performance in the gray-box and white-box setting on CIFAR-10 and ImageNet databases.
\end{itemize}

\section{Related Work}

\textbf{Transformation-based defenses} use one or more transformations to counter adversarial patterns in input examples. For example, \cite{DBLP:conf/iclr/KurakinGB17a} use JPEG compression, and \cite{DBLP:conf/iccv/LiL17} use median-filtering. One of the most effective ways to improve the robustness of transformation-based defense methods is to introduce randomness. The method proposed in \cite{DBLP:conf/icml/YangZXK19} randomly deletes pixels from an input example and reconstructs the deleted pixels using matrix estimation to suppress adversarial patterns. A more recent approach \cite{DBLP:conf/cvpr/RaffSFM19} employs a set of random transformations to defend against strong adversarial attacks generated by a potential adversary with reasonable compute power. 
The method described in \cite{DBLP:conf/ndss/Xu0Q18} applies median smoothing, non-local mean, and bit quantization to an input example and measures the changes in the model's prediction to determine if the input example is adversarial. The approach presented in \cite{DBLP:conf/aaai/TianYC18} uses random rotation and scaling operations to improve adversarial example detection accuracy in white-box setting. Compared to these approaches that use multiple transformations, the proposed method simply applies a single random transformation on the input example to determine if it is adversarial.

\textbf{Neighbor-based defenses} exploit peer wisdom to characterize adversarial examples. Most neighbor-based methods can be categorized into either class-independent or class-conditional methods. Class-independent approaches select K-nearest benign examples of an adversarial example from a reference database and use majority class voting \cite{DBLP:journals/corr/abs-1803-04765}, pixel-wise relations \cite{DBLP:conf/iclr/SvobodaMMBG19} or local intrinsic dimension \cite{DBLP:conf/iclr/Ma0WEWSSHB18} to detect adversarial examples. On the other hand, class-dependent approaches use multiple examples of the predicted class (usually greater than ten) to estimate the kernel density \cite{DBLP:journals/corr/FeinmanCSG17} or construct a valid data distribution \cite{DBLP:conf/nips/LeeLLS18}. An input example that is an outlier with respect to the estimated distribution is labelled adversarial.

\textbf{Prototype-based methods} use a few examples, called prototypes, distilled from the training or reference database for classification. This approach is useful in the few shot learning setting, especially when the number of examples in the reference database is limited. It has been shown that measurement of similarities between test examples and prototypes is quite effective in understanding a model's behavior. Snell \etal \cite{DBLP:conf/nips/SnellSZ17} select prototype examples for each class, and use the nearest-class prototype for classification. Arik and Pfister \cite{DBLP:journals/corr/abs-1902-06292} use prototypes to detect out-of-distribution examples, and provide efficient confidence metrics. Inspired by these methods, we use a prototype from the predicted class of an input example to determine if the input example belongs to the predicted class, and consequently if the input example is adversarial.

\section{Adversarial Example Detection}

\subsection{Problem Definition}

We demonstrate the effectiveness of the proposed adversarial example detection method in the multi-class image classification setting. Given a labeled training set $\mathbb{D}$ containing $N$ examples and $C$ classes such that $\mathbb{D}=\{(\img_1,\la_1),...,(\img_N,\la_N)\}$, with labels $y \in \mathbb{Z}_C$, a classifier ($\eg$, deep neural network) $\net$ is trained on $\mathbb{D}$ to classify an input example (image) $\img$ into one of $C$ classes: $\net(\img)\rightarrow\mathbb{Z}_C$. The loss function for the classifier $f$ is $\phi(\net(\img),y)$. The adversary aims to create an adversarial example $\img'$ that maximizes $\phi$ by using a query example $\img^q$. An additional constraint for the adversary is that the distance between $\img'$ and $\img^q$ should be less than $\epsilon$:
 
\begin{equation}
\img' = \mathop{\mathrm{arg\,max}}_{d(\img^q,\img')<\epsilon} \phi(\net(\img^q),y),
\label{eq:1}
\end{equation}
with $l_{\infty}$ as the assumed distance metric $d$, which provides the adversary maximum flexibility to craft adversarial examples. Furthermore, the adversary's goal is \textit{misclassification} of the crafted adversarial example $\img'$ \ie$\>$ $\net(\img')\neq\net(\img^q)$.

To secure $\net$ from adversarial examples, our goal is to design an adversarial example detector \mbox{$\tau(\img^q)\rightarrow[0,1]$}, such that $\tau()$ outputs a score that indicates whether a query example $\img^q$ is adversarial or benign.

\subsection{Transformation-based Detection}
Let $\star$ denote a generic image transformation operator, and $\textbf{I}^q \star \textbf{T}_l$ represent the application of an image transformation $\textbf{T}_l$ on $\textbf{I}^q$, and result in a perturbed (transformed) image: $\textbf{I}^p_l  = \textbf{I}^q \star \textbf{T}_l$. Transformation $\textbf{T}_l$ (\eg, median filtering, scaling, rotation, Gaussian filtering) where $l\in\{1:L\}$ denotes the $l^{th}$ transformation used in the adversarial example detection method. 
Let $\net^m(\cdot)$ indicate the vectorized output of layer $m$ of deep network $\net$.  The transformation-based detection methods in \cite{DBLP:conf/ndss/Xu0Q18,DBLP:journals/corr/abs-1805-05010,DBLP:conf/nips/HuYGCW19} can be represented as $\tau(\img^q) = \pool(\{r_0, r_1,...,r_L\})$, where the result of feature variations $r_l$ is defined as $r_l=||\net^m(\textbf{I}^p_l) - \net^m(\textbf{I}^q)||_1$. The method $\pool$ aggregates the distances between features of the transformed input examples and the original input example and compares it with a predefined threshold to determine if $\textbf{I}^q$ is adversarial. Because $\textbf{T}_l$ is random, it introduces uncertainties that ideally should be compensated by averaging over a large number or type of transformations. 

\subsection{Proposed Method}
To detect adversarial examples with high accuracy using a small number of transformations, the use of a significantly discriminative characteristic of the adversarial space is important. To this end, instead of introducing complex transformations in $\textbf{T}_l$ that further increase computational complexity, the proposed method is based on how an attacker typically generates an adversarial example.

\begin{figure}[t]%
\centering
\subfloat[$r_l$ distribution]{%
\includegraphics[width=0.47\linewidth, height=1.3in]{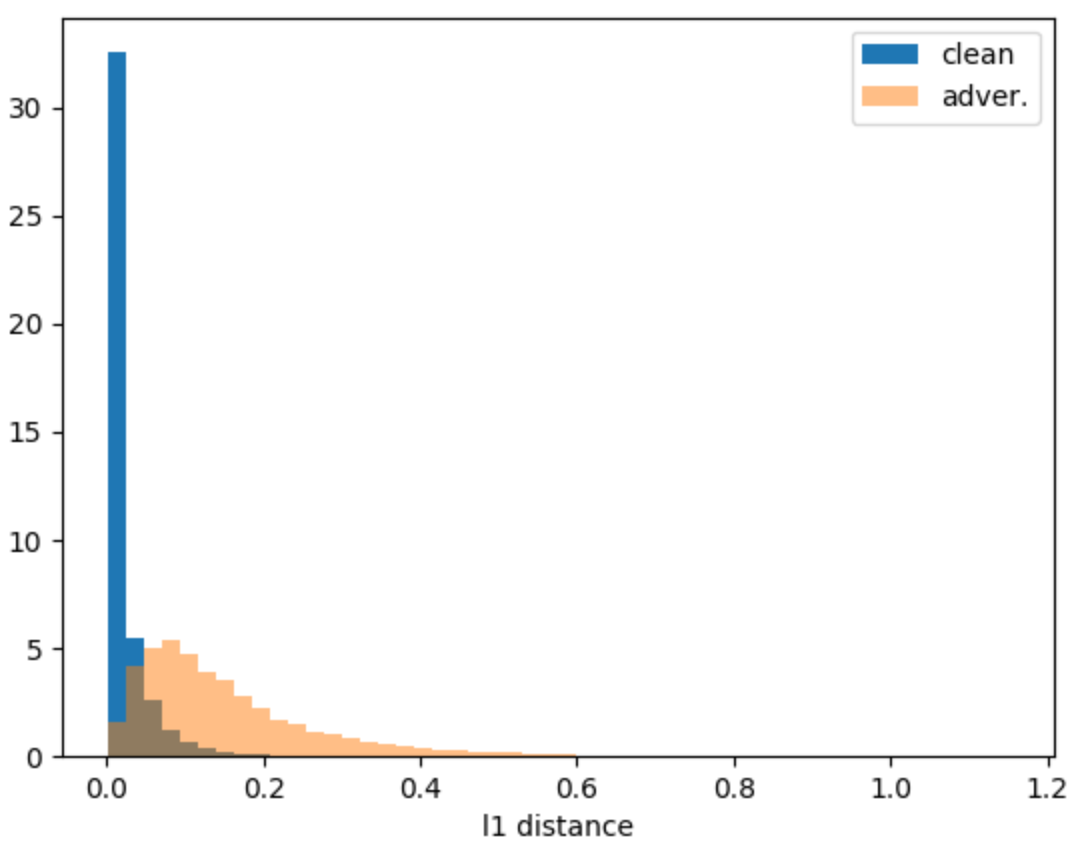}}
\quad
\subfloat[$\alpha_a$ distribution]{%
\centering
\includegraphics[width=0.47\linewidth, height=1.3in]{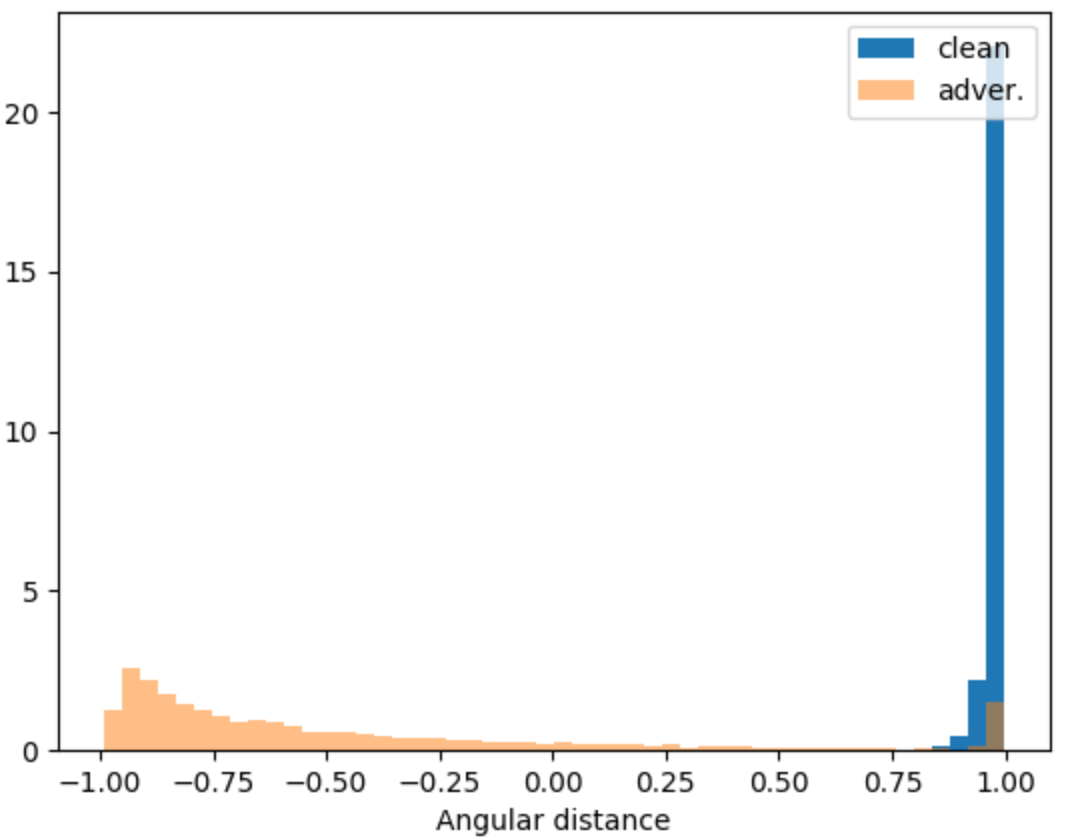}}%
\caption{Distributions of (a) the traditional feature variation score $r_l$ and (b) the proposed score $\alpha_a$ for benign and adversarial examples (generated using FGSM attack) computed for ImageNet database in identical experimental settings.}
\label{f2}
\end{figure}

\textbf{Adversarial gradient direction}: When gradient-based attacks (\eg, FGSM \cite{DBLP:journals/corr/SzegedyZSBEGF13} and PGD \cite{DBLP:conf/iclr/MadryMSTV18}) use an input example $\textbf{I}$ to generate an adversarial example to cause mis-classification, they optimize \mbox{Eq. 1} using gradients computed with respect to $\textbf{I}$:
\begin{equation}
\textbf{I}_{t} = \textbf{I} + \varepsilon_0 sign (\triangledown_{\textbf{I}}  \phi(\textbf{I}, a))
\label{eq:02}
\end{equation}
Here, $a$ is the predicted class label, and $\varepsilon_0$ is the step size of the gradient update. In each iteration, gradient direction $\triangledown_{\textbf{I}}  \phi(\textbf{I}, a)$, abbreviated as $\triangledown \phi(\textbf{I}, a)$, plays a key role in determining the generated adversarial example. The impact of $\triangledown \phi(\textbf{I}, a)$ on vectorized layer output $f^m(\cdot)$ can be quantified by measuring:
\begin{equation}
\Delta f^m(\textbf{I},a) =  f^m(\textbf{I}_{t}, a) -  f^m(\textbf{I}, a)
\label{eq:20}
\end{equation}
We refer to the direction of the vector $\Delta f^m(\textbf{I},a)$ as \textbf{Adversarial Gradient Direction (AGD)} targeted at class $a$. We observe that the direction of $\Delta f^m(\textbf{I},a)$ changes significantly if $\textbf{I}$ corresponds to adversarial example $\textbf{I}'$. However, the direction remains consistent when $\textbf{I}$ corresponds to a benign example. We exploit this property of AGD to detect adversarial examples:

Given a query image $\textbf{I}^q$, we compute the following score:
\begin{equation}
\alpha_a = <\Delta f^m(\textbf{I}^q, a),  \Delta f^m(\textbf{I}^p,a)>
\label{eq:211} 
\end{equation}
In Eq. \ref{eq:211}, \mbox{$<\cdot,\cdot>$} computes the angular similarity between two vectors. Fig. \ref{f2}(a) and \ref{f2}(b) respectively show the distributions of the traditional feature variation score $r_l$ and the proposed score $\alpha_a$ under the same random pixel perturbation $\bm{\mu}$. Note the significantly less overlap between adversarial and benign example distributions of $\alpha_a$ compared to corresponding distributions of $r_l$.

To improve the overall detection accuracy, AGDs can be calculated not only for the predicted class $a$ with the highest probability but for each class in the training set. One suggested strategy is to compute AGDs for the top $K$ classes that yield the highest probability, and choose $K$ based on the trade-off between the desired performance and required computation. Eq. \ref{eq:211} is thus extended as follows:

\begin{equation}
\alpha_k = <\Delta f^m(\textbf{I}^q, k),  \Delta f^m(\textbf{I}^p,k)>
\label{eq:200} 
\end{equation}

The detection performance is found to saturate after \mbox{$K=4$} (see Table \ref{T2}) for both CIFAR-10 and ImageNet databases.

\begin{figure}[t]%
\subfloat[$\textbf{I}$]{
\centering
\includegraphics[width=0.3\linewidth, height=0.9in]{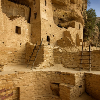}}
\centering
\subfloat[$\textbf{I}^p$]{
\includegraphics[width=0.3\linewidth, height=0.9in]{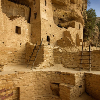}}
\centering
\subfloat[$\textbf{I}^n$]{
\includegraphics[width=0.3\linewidth, height=0.9in]{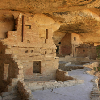}}
\\
\centering
\subfloat[$\textbf{I}'$]{
\includegraphics[width=0.3\linewidth, height=0.9in]{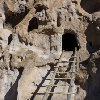}}
\centering
\subfloat[$\textbf{I}'^p$]{
\includegraphics[width=0.3\linewidth, height=0.9in]{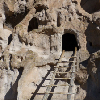}}
\centering
\subfloat[$\textbf{I}'^n$]{
\includegraphics[width=0.3\linewidth, height=0.9in]{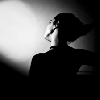}}
\caption{Sample benign and adversarial examples (a) and (d) for ImageNet database. (b) and (e) show the corresponding perturbed (transformed) examples, and (c) and (f) show the corresponding nearest neighbors from the predicted class for (a) and (d).}
\label{f1100}
\end{figure}

\textbf{Use of predicted class prototype}:
Because the perturbation applied on an input example is random, a transformed example may not always exhibit the desired property of AGDs used for adversarial example detection. In other words, a transformed example created from an adversarial example may have similar AGD to the adversarial example, while it may have different AGD compared to its benign neighbors.

To handle this anomaly, we use a prototype benign example $\textbf{I}^n$ that belongs to the predicted class $a$ from a reference database $\mathbb{D}'$.
\begin{figure}%

\subfloat[$\beta_a$ distribution]{%
\centering
\includegraphics[width=0.5\linewidth, height=1.3in]{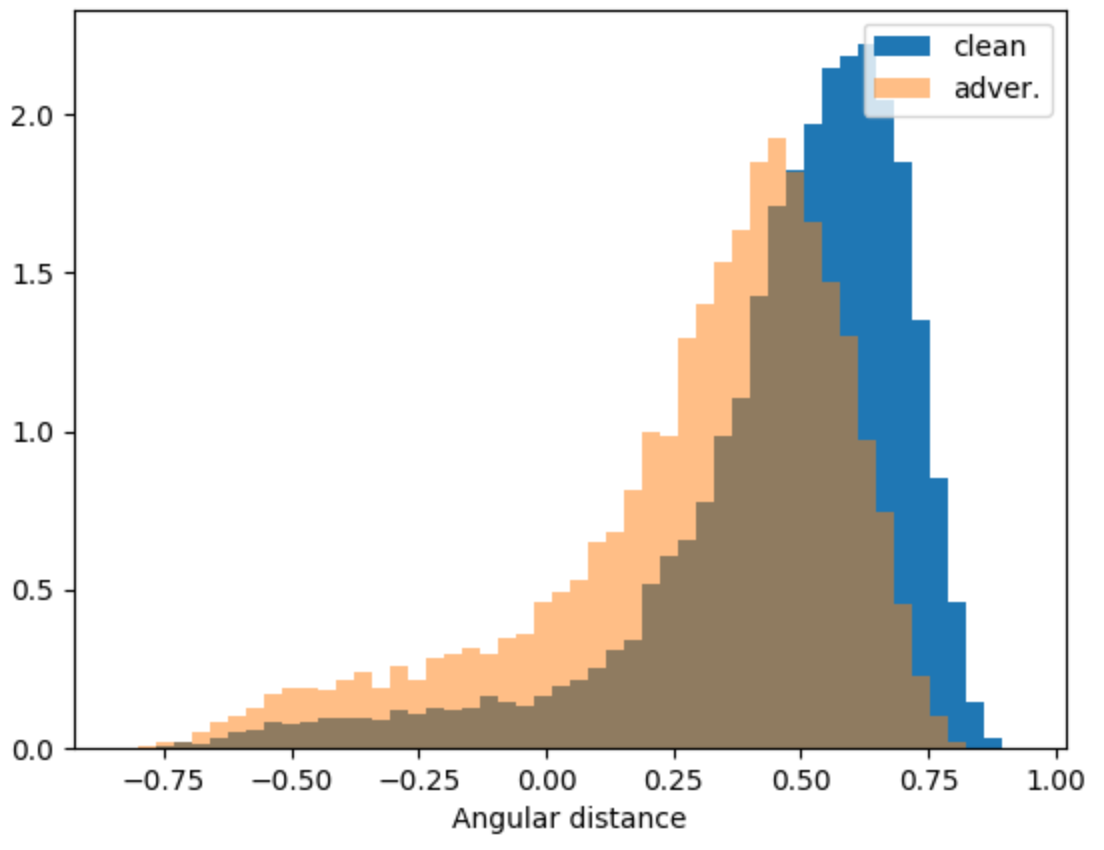}}%
\centering
\subfloat[$\gamma_a$ distribution]{%
\includegraphics[width=0.5\linewidth, height=1.3in]{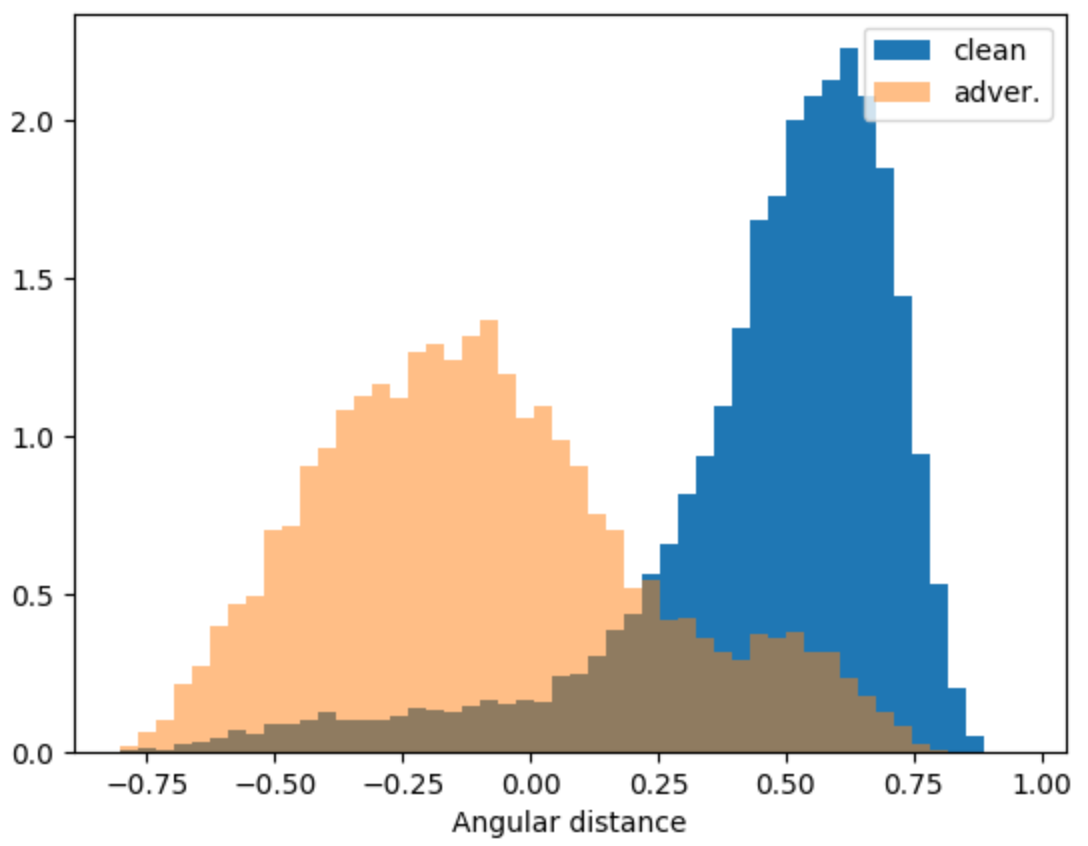}}
\\
\centering
\subfloat[$\beta_{c_2}$ distribution]{%
\includegraphics[width=0.5\linewidth, height=1.3in]{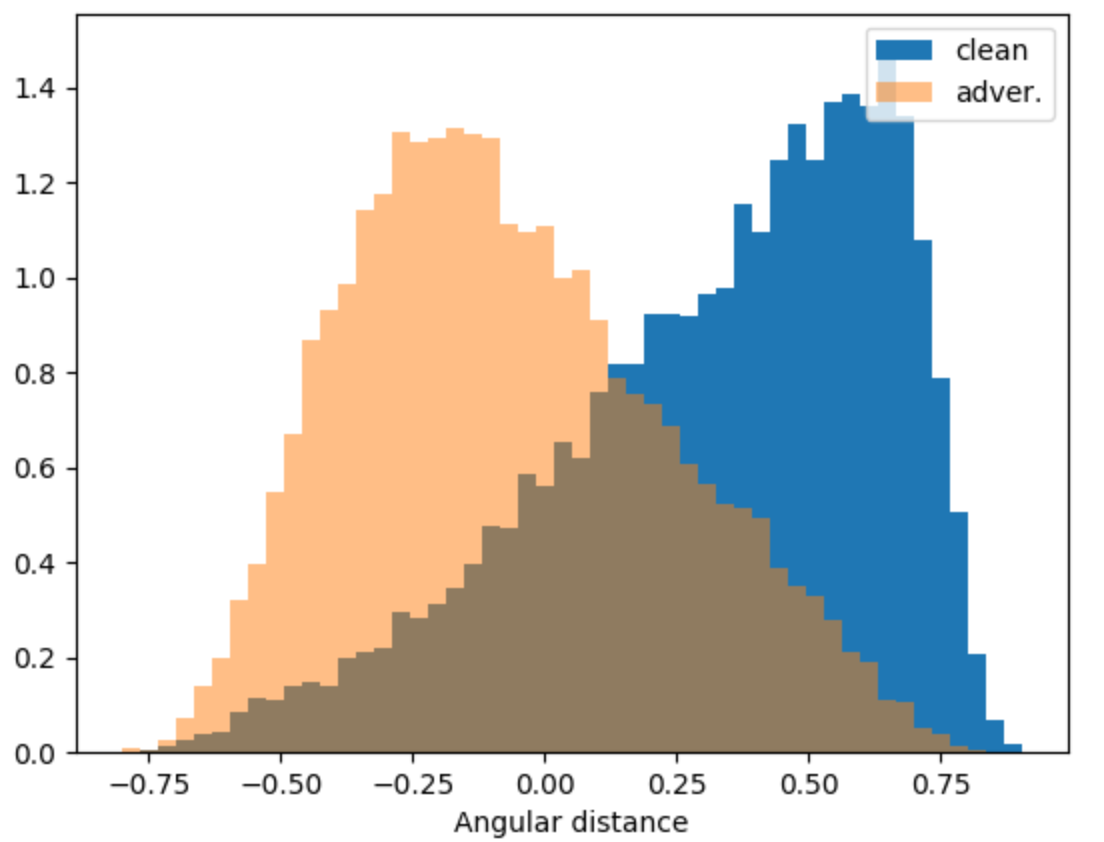}}
\centering
\subfloat[$\gamma_{c_2}$ distribution]{%
\includegraphics[width=0.5\linewidth, height=1.3in]{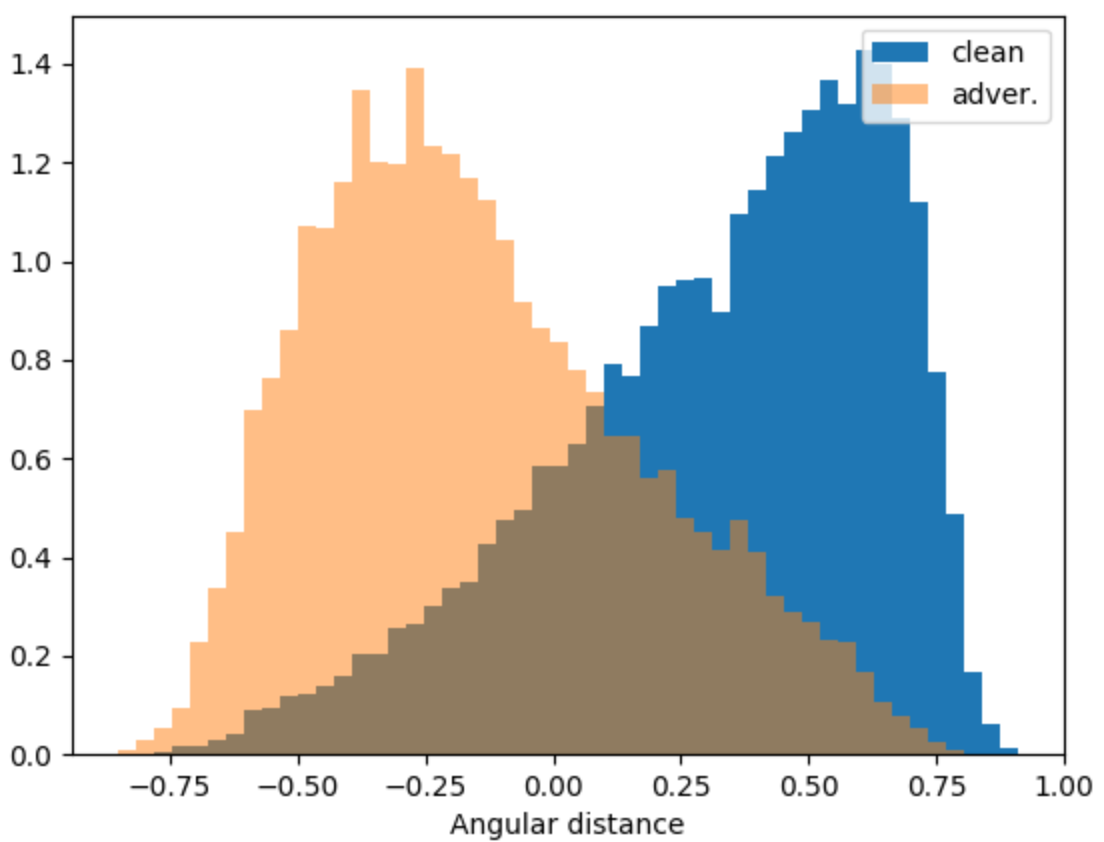}}

\caption{$\>$Distributions of $\beta_a, \gamma_a, \beta_{c_2}$, and $\gamma_{c_2}$ for benign and adversarial examples computed for ImageNet database using the same random transformation. }
\label{f22}
\end{figure}
Recall that AGD of $\textbf{I}^q$ is jointly determined by target class $k$ and $f^m(\textbf{I}^q)$. Similarly, let $\Delta f^m(\textbf{I}^n,k)$ correspond to the prototype benign example $\textbf{I}^n$ from class $a$ that is sufficiently close to $\textbf{I}^q$ on the manifold. In this work, the nearest neighbor of the query example in the reference database $\mathbb{D}'$ based on the embedding distance is selected as the prototype.

To capture the relationship between AGDs of $\textbf{I}^q$ and $\textbf{I}^n$, we define a score $\beta_k$:
\begin{equation}
\beta_k = <\Delta f^m(\textbf{I}^q , k),  \Delta f^m(\textbf{I}^n,k)>,
\label{eq:2} 
\end{equation}
such that $\beta_k$ should be high if $\textbf{I}^q = \textbf{I}$ but low if $\textbf{I}^q = \textbf{I}'$. This is because an adversarial example is typically crafted by applying perturbation on a query example in the direction of a decision plane relative to the query example. Therefore, if two examples are neighbors in the feature space, their corresponding directions towards a decision plane (defined by $k$) should be similar. Specifically, because $\textbf{I}^n$ is a benign example from class $a$, AGD of $\textbf{I}^n$ should act a prototype for $\textbf{I}^q$ and encapsulates the AGD of the local manifold. However, a crafted adversarial example intrinsically belongs to a different class. Therefore, the gradient update path used by the adversarial example generation process is different.
 
\begin{figure*}%

 \centering
\subfloat[Sampled pos.]{%
\includegraphics[width=0.12\linewidth, height=1.55in]{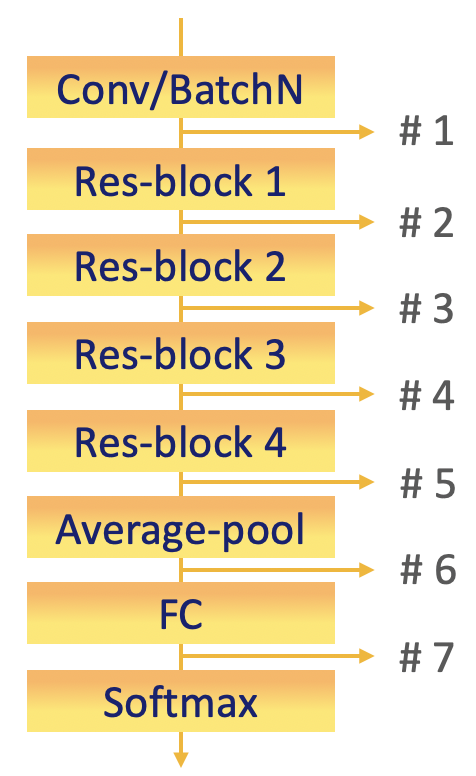}}%
\quad
 \centering
\subfloat[Discriminatory power of $\alpha$]{%
\includegraphics[width=0.26\linewidth, height=1.55in]{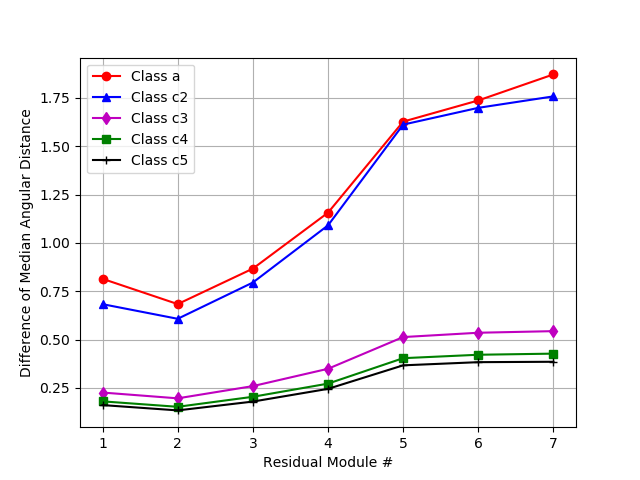}}%
\quad
 \centering
\subfloat[Discriminatory power of $\beta$]{%
\includegraphics[width=0.26\linewidth, height=1.55in]{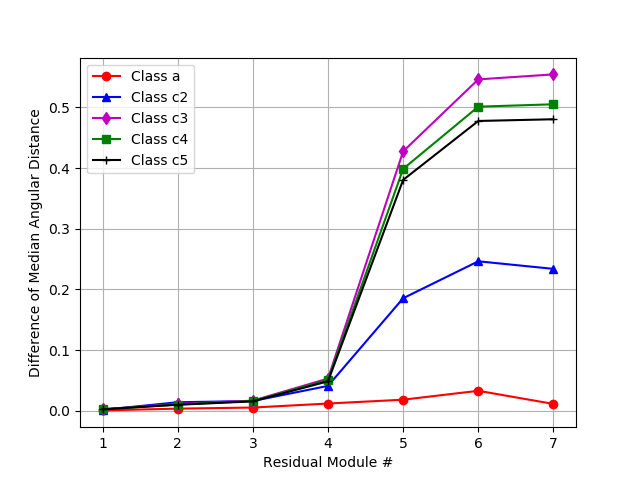}}%
\quad
 \centering
\subfloat[Discriminatory power of $\gamma$]{%
\includegraphics[width=0.26\linewidth, height=1.55in]{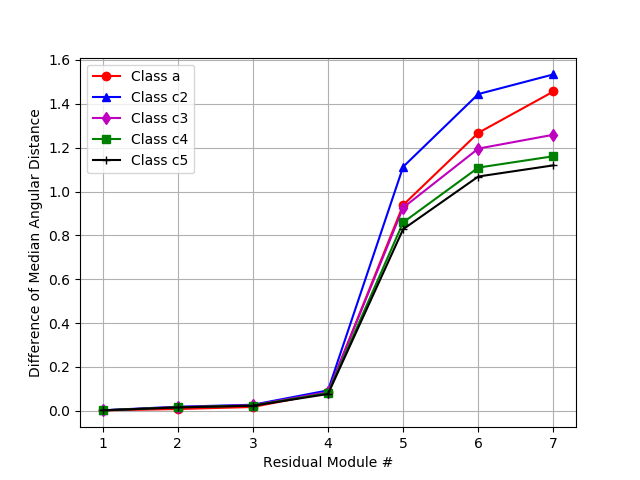}}%

\caption{Discriminatory power of (b) $\alpha$, (c) $\beta$, and (d) $\gamma$ measured using difference of median angular distances between benign and adversarial examples at different ResNet-18 network layers $m$ shown in (a). (Best viewed in color)}
\label{f4}%
\end{figure*}

It is straightforward to derive the third score $\gamma_k$ which measures the similarity between AGD of the transformed example $\textbf{I}^p$ and $\textbf{I}^n$:
\begin{equation}
\gamma_k = <\Delta f^m(\textbf{I}^p , k),  \Delta f^m(\textbf{I}^n, k)>
\label{eq:3} 
\end{equation}
Because $f^m(\cdot)$ is trained to be robust to small perturbations in benign input examples and enforce low intra-class variability, it is expected that $f^m(\textbf{I}^p)$ and $f^m(\textbf{I}^n)$, and their corresponding AGDs will be similar to each other if $\textbf{I}^q=\textbf{I}$, and quite different if $\textbf{I}^q = \textbf{I}'$.

 Figure \ref{f1100} shows benign and adversarial examples and their corresponding neighbors from ImageNet database. The distributions of $\beta_a$ and $\gamma_a$ are shown in \mbox{Fig. \ref{f22}}. Also visualized are the distributions of $\beta_{c_2}$ and $\gamma_{c_2}$ when AGDs are computed for the 2$nd$ most probable class (noted as subscript $c_2$). Note that the distributions of $\beta_{c_2}, \gamma_{a}$, $\gamma_{c_2}$ have significantly less overlap compared to $\beta_{a}$. This is due to overfitting of the adversarial example to target class $a$ resulting in an adversarial example exhibiting similar characteristic AGD properties as its nearest neighbor from class $a$.

\textbf{Detection using adversarial gradient directions:}
Thus far, we discussed the key elements ($\alpha_a ,\beta_a$, and $\gamma_a$) of the proposed adversarial example detector. The performance of the detector primarily depends on the following two observed properties of AGDs:
\\
(i) \textbf{Transformation consistency}: AGDs of adversarial examples change quite significantly after a single random transformation compared to AGDs of benign examples.
\\
(ii) \textbf{Neighborhood smoothness}: Benign examples and the nearest neighbors that belong to the same class share similar AGDs, while adversarial examples share different AGDs with their benign \mbox{neighbors} from the predicted class.

The proposed method is summarized in Algorithm \ref{AGD}. The output $\tau$ is obtained using the trained classifier  $\pool^*\{\cdot\}$: 
\begin{equation}
\tau(\img^q) = \pool^*(\{\alpha_a,\beta_a,\gamma_a,\alpha_{c_1},\beta_{c_1},\gamma_{c_1}...,\alpha_K,\beta_K,\gamma_K \})
\label{eq:4} 
\end{equation}

\textbf{Computational complexity:} Instead of $L$ transformations, the proposed method uses a single transformation. Furthermore, the number of classes used in our method is a fixed parameter $K<<N$; $N$ is the total number of classes. Compared to state-of-the-art methods (\eg, \cite{DBLP:conf/icml/RothKH19}) that use a large $L$ and $N$ and have computational complexity of $O(LN)$, the computational complexity of the proposed method is $O(K)$.

\section{Experimental Evaluation}
The proposed method is evaluated as a defense mechanism for four widely used deep network architectures, ResNet-18, ResNet-50 \cite{DBLP:conf/cvpr/HeZRS16}, GoogleNet \cite{DBLP:conf/cvpr/SzegedyVISW16}, and DenseNet \cite{DBLP:conf/cvpr/HuangLMW17}, against five different state-of-the-art attacks including three gradient-based attacks: FGSM \cite{DBLP:journals/corr/SzegedyZSBEGF13}, CW \cite{DBLP:conf/sp/Carlini017}, PGD \cite{DBLP:conf/iclr/MadryMSTV18}, a decision-based attack: Boundary \cite{DBLP:conf/iclr/BrendelRB18}, as well as an image deformation-based attack: Adef \cite{DBLP:conf/iclr/AlaifariAG19}. Experiments are conducted on two different databases, CIFAR-10 \cite{CIFAR10} and ImageNet \cite{DBLP:journals/cacm/KrizhevskySH17}. In experiments conducted on CIFAR-10 database, $20,000$ images are used to train the baseline residual network, $20,000$ are used as reference data, $10,000$ are used to generate adversarial examples and learn the parameters of the proposed method, and the remaining $10,000$ testing images are used for evaluation. The experiments on ImageNet database follow the protocol in \cite{DBLP:journals/cacm/KrizhevskySH17}. The entire validation set of ImageNet containing $50,000$ images is used. $10,000$ images are used for neighborhood retrieval, $20,000$ to train the model parameters, and the remaining $20,000$ for testing. We used the pre-trained network parameters for ResNet-50 and DenseNet-121 on ImageNet. The proposed method is compared with state-of-the-art adversarial example detection methods that are transformation-based (Feature Squeeze \cite{DBLP:conf/ndss/Xu0Q18} and \cite{DBLP:conf/cvpr/SzegedyVISW16}), and nearest neighbor-based (Mahalanobis \cite{DBLP:conf/nips/LeeLLS18} and DkNN \cite{DBLP:journals/corr/abs-1803-04765}). Each method is trained and tested on identical training and testing set partitions so that the reported results are comparable.

\begin{figure*}%
\centering
\subfloat[]{%
\centering
\includegraphics[width=0.23\linewidth, height=1.2in]{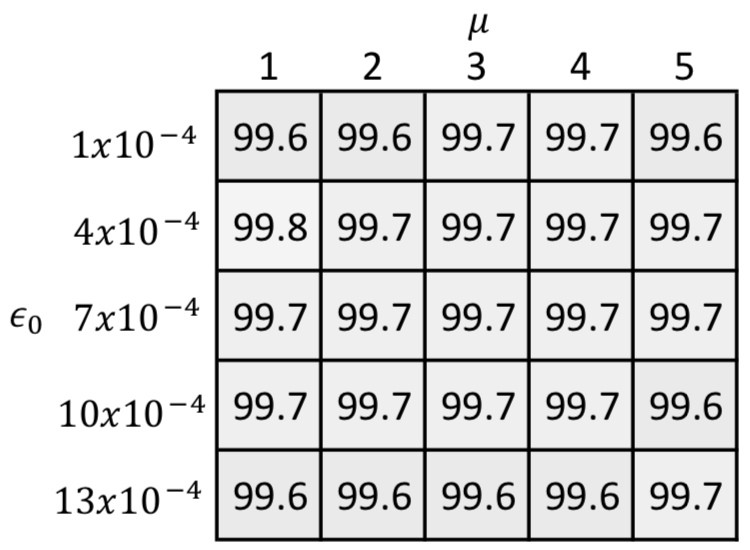}}%
\subfloat[]{%
\centering
\includegraphics[width=0.23\linewidth, height=1.2in]{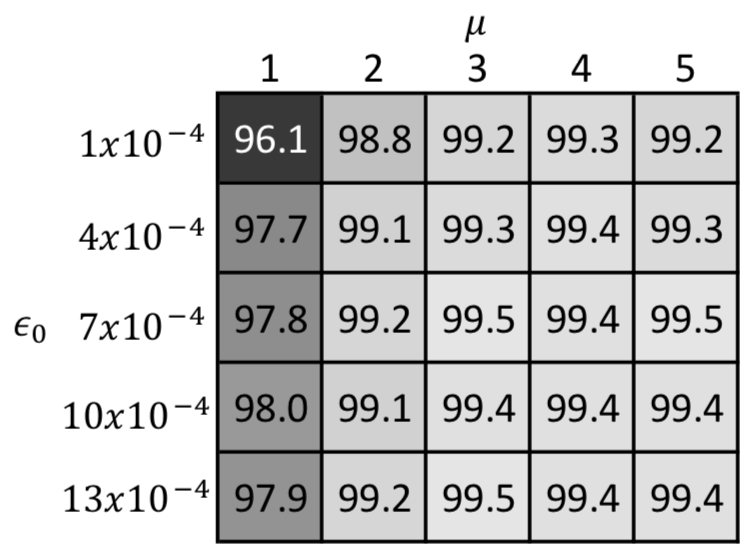}}%
\subfloat[]{%
\centering
\includegraphics[width=0.35\linewidth, height=1.2in]{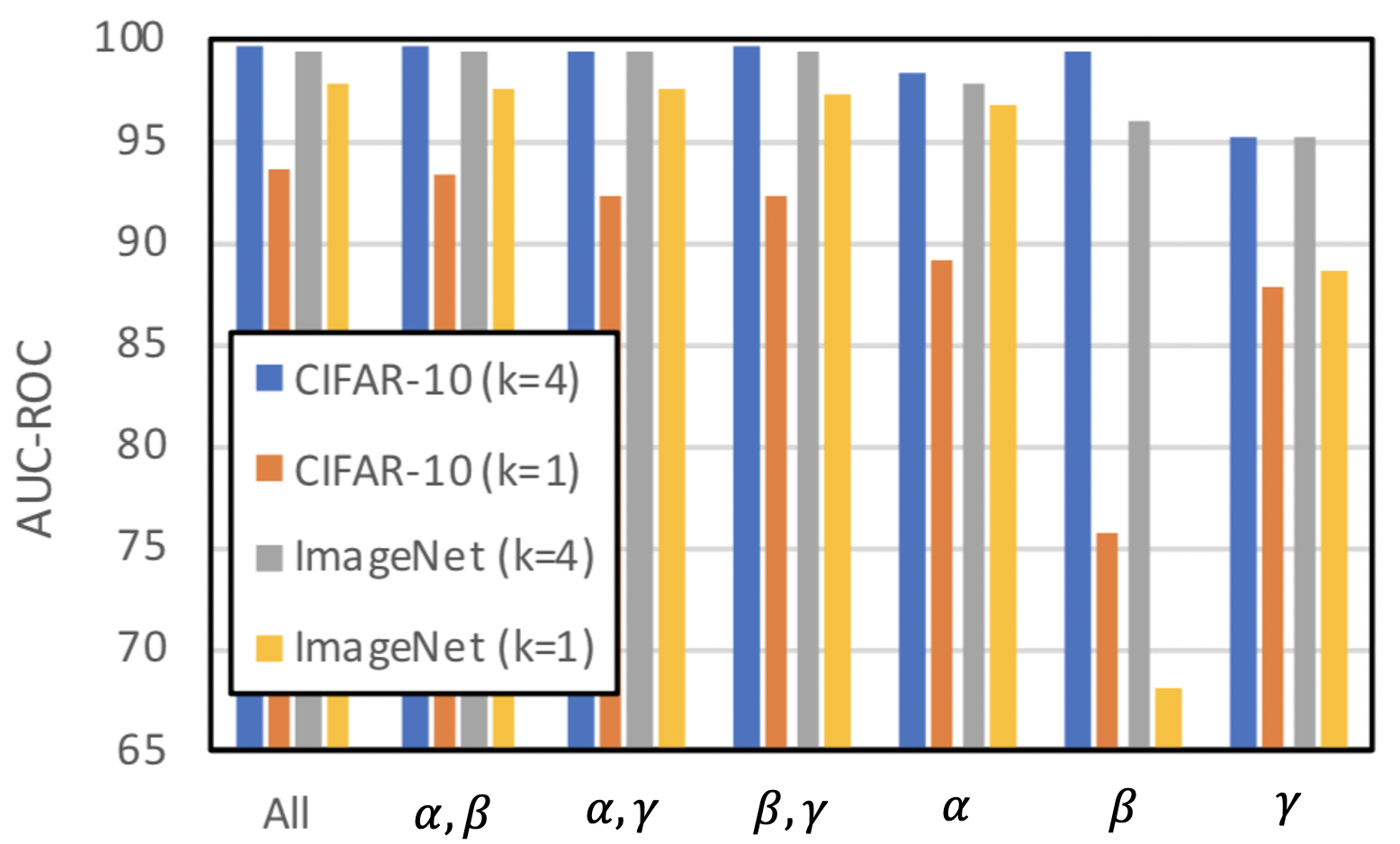}}%
\caption{Adversarial example detection performance (AUC-ROC (\%)) on (a) CIFAR-10 and (b) ImageNet database for different combinations of random perturbation $\mu$ and step size $\epsilon_0$. Shown in (c) is the detection performance on the two databases for different combinations of the proposed angular similarity scores $\alpha, \beta$, and $\gamma$ for predicted classes under $K=1$ and $K=4$.
}
\label{f55}

\end{figure*}

\textbf{Experimental settings}: Adversarial examples are generated under the $l_{\infty}$ norm constraint. The maximum permitted perturbation is set to 8 and 16 pixels, respectively, for CIFAR-10 and ImageNet databases. The step size $\varepsilon_0$  is fixed to 0.0013. A single random perturbation $\mu$ is permitted on a maximum of three pixels for input transformation. A random forest classifier (30 trees) is used as $\pool^*\{\cdot\}$. Area Under the Curve (AUC) of the Receiver Operating Characteristics (ROC) curve is used as the evaluation metric. Following the protocol in \cite{DBLP:conf/nips/LeeLLS18}, examples that are: (i) classified correctly, and (ii) can be perturbed to successfully attack the deep network, are used for training and testing.

\begin{algorithm}
	\KwIn{$\textbf{I}^q$, $f(\cdot)$, $m$, $\textbf{T}$, $K$, $\epsilon_0$, $\mathbb{D}'$, and $\pool^*()$}
	\KwOut{  $\tau(\img^q)$ } 
	  Compute the predicted class $a = f(\textbf{I}^q)$ \;
	  Save top $K$ class indices to $\mathbb{C} = \{a, c_2, ... c_k\}$   \;
    Retrieve $\textbf{I}^n$ of $\textbf{I}^q$ from $\mathbb{D}'$ based on $a$ \;
    Obtain a transformed image $\textbf{I}^p$ with $\textbf{I}^p = \textbf{I}^q \star \textbf{T}$ \;
    Initialize an empty feature vector $\textbf{v} = \{\}$  \;
	\For  {$c_k \in \mathbb{C}$} {
	    Run FGSM: $\textbf{I}^q_k = \textbf{I}^q +\varepsilon_0 sign(\triangledown \phi(\textbf{I}^q, c_k))$ \;
	  	Run FGSM: $\textbf{I}^p_k = \textbf{I}^p +\varepsilon_0 sign(\triangledown \phi(\textbf{I}^p, c_k))$ \;
	    Run FGSM: $\textbf{I}^n_k = \textbf{I}^n +\varepsilon_0 sign(\triangledown \phi(\textbf{I}^n, c_k))$ \;
	    $\Delta f^m(\textbf{I}^q_k, c_k) = f^m(\textbf{I}^q_k) -  f^m(\textbf{I}^q) $ \;
	    $\Delta f^m(\textbf{I}^p_k, c_k) = f^m(\textbf{I}^p_k) -  f^m(\textbf{I}^p)$ \;
	    $\Delta f^m(\textbf{I}^n_k,c_k) = f^m(\textbf{I}^n_k) -  f^m(\textbf{I}^n) $ \;
	    $\alpha_k = <\Delta f^m(\textbf{I}^q_k, c_k ),  \Delta f^m(\textbf{I}^p_k,c_k) >$ \;
	    $\beta_k = <\Delta f^m(\textbf{I}^q_k , c_k ),  \Delta f^m(\textbf{I}^n_k,c_k )>$ \;

	    $\gamma_k = <\Delta f^m(\textbf{I}^p_k , c_k ),  \Delta f^m(\textbf{I}^n_k,c_k )>$ \;
	    $\textbf{v}$ = $\textbf{v} \cup \{\alpha_k, \beta_k, \gamma_k\} $ }
	$\tau(\img^q) = \pool^*(\textbf{v})$.
	\caption{Detection using AGDs}
	\label{AGD}
\end{algorithm}

\subsection{Parameter analysis}

\textbf{Network layers used for detection:}
The goal of this experiment is to determine the deep network layer(s) $m$ that are useful for computing AGDs. Experiments are conducted on CIFAR-10 database using the ResNet-18 architecture and ImageNet database using DenseNet-121 architecture to detect the FGSM attack. The difference between median angular similarities of AGDs, $\alpha_k, \beta_k$ and $\gamma_k$, of benign and adversarial example distributions are computed after each residual module in the ResNet-18 pipeline (see Fig. \ref{f4}). It is observed that adversarial and benign example distributions are comparatively separable after residual block $\#5$. Given this, $\alpha_k, \beta_k$, and $\gamma_k$ are computed for the embedding (layer output $\#6$) and logit layer (layer output $\#7$), and combined for the final decision. Note that the discriminatory power of $\alpha_k$ is maximal for the top two most confident classes, while that of $\beta_k$ and $\gamma_k$ is maximal for the 3$rd$ and 2$nd$ most confident class, respectively. This observation indicates that it is necessary to take into account the results of AGDs for multiple classes to improve the overall detection performance.

\textbf{AGD parameters:}
The objective of this experiment is to determine the optimal combination of parameters that maximizes the overall detection performance. The parameters include (i) number of predicted classes $K$, (ii) added random perturbation $\mu$, and (iii) the step size $\epsilon_0$. A parameter search is performed on the parameter space with $K$ ranging from one to five, and $\mu$ and $\epsilon_0$ as shown in Fig. \ref{f55}. Parameter values that minimize the required perturbation on an input example (for imperceptibility) are considered optimal. 

\begin{table}[h]\centering
\scalebox{0.9}{
\begin{tabular}{cccccccc}
\Xhline{\arrayrulewidth}

Class number ($K$) & 1 & 2 & 3 & 4 & 5\\
\hline
CIFAR-10    & 94.1  & 97.5  & 99.7 &  \textbf{99.8}  &  \textbf{99.8}     \\
ImageNet    & 98.2  & 99.2  & 99.4 &  \textbf{99.5}  &  \textbf{99.5}     \\
\hline
\Xhline{\arrayrulewidth}
\end{tabular}}
\caption{Adversarial example detection performance (AUC-ROC in \%) for top $K$ predicted classes.}
\label{T20}
\end{table}
Table \ref{T20} shows the detection performance for different values of $K$. The performance saturates after $K=4$. For $K=4$, the best performing combinations of $\mu$ and $\epsilon_0$ are shown in Figs. \ref{f55} (a) and \ref{f55} (b). 

\textbf{Number of random transformations:}
To measure the impact of the number of random transformations on our method, different number of random transformations (upto five) are used to detect the FGSM attack. The angular similarity scores for the selected transformations are concatenated and fed to the classifier in Eq. 8.

\begin{table}[H]\centering
\scalebox{0.9}{
\begin{tabular}{cccccccc}
\Xhline{\arrayrulewidth}
Random trans. number & 1 & 2 & 3 & 4 & 5\\
\hline
CIFAR-10    & 99.6  & \textbf{99.7}  & \textbf{99.7} &  \textbf{99.7}  &  \textbf{99.7}     \\
ImageNet    & 99.5  & 99.5  & 99.6 &  \textbf{99.7}  &  \textbf{99.7}     \\
\hline
\Xhline{\arrayrulewidth}
\end{tabular}}
\caption{Adversarial example detection performance (AUC-ROC in \%) for different number of random transformations.}
\label{T201}
\end{table}
Table \ref{T201} shows that only a slight increase in detection performance is obtained using more than one random transformation. Hence, a single random transformation in used in other experiments.

\subsection{Ablation study}
The objective of this experiment is to determine the impact of each angular similarity score $\alpha_k, \beta_k$, and $\gamma_k$ on the final performance. Experiments are conducted for top $K$ predicted classes, $K=1$ and $K=4$, and the obtained performance for different score combinations is visualized in Fig.~\ref{f55} (c).
\begin{table*}\centering
\scalebox{0.9}{
\begin{tabular}{cc|ccccc|cccc}
\Xhline{\arrayrulewidth}

Network (Dataset)         & Method & FGSM    & ADef.& CW &  PGD  & Boundary& ADef. &  CW  & PGD &  Boundary  \\
  \hline
  
                & Rand-1  & 78.8    & 82.8 & 88.8 & 92.0 &74.3 &78.7 & 88.7 & 92.0 & 74.3 \\
                & Median  & 91.7    & 86.5 & 91.6 & 93.0 &92.2 &86.4 & 91.6 & 93.0 & 92.1 \\
{ResNet-18}     & MA.     & 96.8    & 94.0 & 94.1 & 96.4 &98.3 &\textbf{93.9} & \textbf{94.1} & 96.4 & 98.4     \\
(CIFAR-10)       & DkNN    & 99.2    & 93.7 & 93.4 & 99.3 &99.5 &93.7 & 93.4 & 99.3 & \textbf{99.5}      \\

                & Ours    & \textbf{99.6}    & \textbf{94.7} & \textbf{96.4} & \textbf{99.7} &\textbf{99.6} &90.3 & 93.1 & \textbf{99.7} & 99.4     \\
 \cline{1-11}
                 & Rand-1 & 58.4    & 53.1 & 52.5 & 51.7  & 55.7 & 58.4 & 52.5 & 51.7 & 55.7 \\
                 & Median & 60.0    & 54.4 & 52.8 & 52.0  & 74.3 & 54.4 & 52.8 & 52.0 & 74.3 \\
{GoogleNet}      & MA.    & 95.8    & 94.1 & 91.4 & 95.1  & 97.9 & 94.1 & 91.4 & 95.1 & 94.8 \\
(CIFAR-10)          & DkNN   & \textbf{99.0}    & 97.1 & 90.4 & \textbf{99.0}  & 99.1 & \textbf{97.1} & 90.4 & \textbf{99.0} & 99.1     \\

                 & Ours   & 98.0  & \textbf{97.2} &\textbf{98.2}& 98.5  &  \textbf{96.6}   &  96.9 & \textbf{98.1}  & 98.4 &  \textbf{92.1}     \\
  \hline
  
  \hline
                & Rand-1  & 50.4 & 51.4 & 54.1 & 69.3 & 50.0 & 50.4& 54.1 & 69.3 & 50.0 \\
                & Median  & 85.7 & {87.2} &  {91.2} & {89.0} & 86.7 & {87.2}& {91.2} & {89.0} & 86.7 \\
{ResNet-50}     & MA.     & 58.4    & 58.0 &58.9& 56.3  &  77.9   &  58.0 &55.8  &56.5 & 77.4     \\
(ImageNet)      & DkNN    & 67.2    & 63.7 &61.9& 62.7  &  81.6   &  63.7 &61.9  &62.7 & 81.6      \\

                & Ours    & \textbf{99.9}    & \textbf{99.9} &\textbf{99.7}& \textbf{99.9}  &  \textbf{99.8}   &  \textbf{99.9} &\textbf{99.6}  & \textbf{99.9}& \textbf{99.8}       \\
 \cline{1-11}
                & Rand-1  & 50.1 & 51.0 & 53.5 & 66.3 & 50.0 & 50.1 & 53.5 & 66.3 & 50.0 \\
                & Median  & 86.2 & 83.3 & {90.0} & 86.7 & 85.5 & 83.3 & \textbf{90.0} & 87.7 & 85.5 \\
{DenseNet-121}  & MA.     & 72.3    & 70.3 &70.8& 71.4  &  70.7   &  71.4 &71.7  &71.4 &  44.3       \\
(ImageNet)      & DkNN    & 63.7    & 60.6 &59.4& 60.3  &  80.0   &  60.6 &59.4  &60.3 &  80.0     \\

                & Ours    & \textbf{99.5}  & \textbf{99.1} &\textbf{91.0}& \textbf{99.5}  &  \textbf{98.1}   &  \textbf{97.4} & 83.7  &\textbf{99.5} &  \textbf{96.1}     \\
  \hline
\Xhline{\arrayrulewidth}
\end{tabular}}
	\caption{Comparison of the proposed adversarial example detection method with state-of-the-art methods on the ImageNet database. Performance is reported using area under curve (AUC) (\%) of the ROC curve. The top performing algorithms are highlighted. (L) Training and testing on the same attack, and (R) training using FGSM attack and testing on other attacks.}
\label{Benchmark}
\end{table*}
Fig. \ref{f55} (c) shows that the performance obtained using any combination of two scores is lower than the performance obtained by aggregating all three scores. Using only one similarity score is not recommended since it may not appropriately capture either of the two AGD properties.

\subsection{Comparison with state-of-the-art}
Comparison of the proposed method with state-of-the-art adversarial example detection methods is shown in \mbox{Table \ref{Benchmark}}. For transformation-based methods, we compare the proposed method to \cite{DBLP:conf/nips/HuYGCW19} and \cite{DBLP:conf/ndss/Xu0Q18}. These two methods use a series of transformations to transform the input example. For a fair comparison with our method, we limit the comparison to two fundamental but most effective transformations. The first is one-time Gaussian random perturbation used in the first stage of \cite{DBLP:conf/nips/HuYGCW19} (denoted as Rand-1)and the second is the 2$\times$2 median filter used in Xu \etal \cite{DBLP:conf/ndss/Xu0Q18} (denoted as Median). The $l_1$ score used in these methods is used for benchmarking. For neighbor-based defenses, we benchmark against DkNN \cite{DBLP:journals/corr/abs-1803-04765} and MA \cite{DBLP:conf/nips/LeeLLS18}. Each of these methods uses over 20 reference examples to model benign example distribution for adversarial example detection. The parameters of each method are optimized to maximize performance on CIFAR-10 and ImageNet databases. The performance is reported in two different settings: (i) training and testing with the same attack algorithm, (ii) training on adversarial samples generated by FGSM and testing on other attacks. The results shown in Table \ref{Benchmark} indicate that the proposed approach significantly outperforms state-of-the-art methods on CIFAR and ImageNet databases. The proposed method also scales reasonably well when the number of classes increase from 10 (CIFAR-10) to 1,000 (ImageNet).

\begin{table}[b]\centering
\scalebox{0.9}{
\begin{tabular}{c|cc|cc}
\Xhline{\arrayrulewidth}

Target  &  \multicolumn{2}{c|}{Attack Success Rate} & \multicolumn{2}{c}{Detection AUC} \\
        & LR=0.1  & LR=0.03 &  LR=0.1 & LR=0.03 \\
\hline
Rand-1        & 79.4  & 78.0  & 53.0 & 53.1  \\
Ours (K=1)    & 74.9  & 70.9  & 53.3 & 60.8  \\
Ours (K=4)    & 53.5  & 60.3  & 64.1 & 62.5  \\
\hline
\Xhline{\arrayrulewidth}
\end{tabular}}
\caption{White-box attack performance (AUC-ROC in \%). LR denotes the learning rate used while conducting the attack.}
\label{T2}
\end{table}

\textbf{White-box attack setting:} In the earlier experiments (gray-box attack setting), it was assumed that the attacker has no knowledge of the adversarial detection approach used. In this experiment, it is assumed that the attacker has complete knowledge of the adversarial detection method. The adaptive attack setting described in \cite{DBLP:conf/nips/HuYGCW19} is used. A subset of 1,000 images from ImageNet database are selected. PGD attack is used to jointly optimize $\lambda\phi(\net(\img),y)$ and maximize all similarity scores in $\{\alpha_a,\beta_a,\gamma_a,\alpha_{c_2},\beta_{c_2},\gamma_{c_2}...,\alpha_k,\beta_k,\gamma_k\}$, simultaneously. The parameter $\lambda$ is set to two. The attack success rate which indicates if an adversarial example is able to fool the deep network, and detection AUC-ROC for $K= \{1,4\}$ are reported in Table \ref{T2}. The performance is compared with Rand-1, where $r=||\net^m(\textbf{I}^p)  - \net^m(\textbf{I}^q)||_1$ and $\lambda\phi(\net(\img),y)$ are jointly minimized. The maximum permitted perturbation $l_{\infty}$ is set to $0.1$ of pixel value range, and the step size for PGD is set to 100. The results for two different learning rates, $0.1$ and $0.03$, are compared. Experimental results show that our method significantly reduces the attack success rate when $\lambda\phi(\net(\img),y)$ and detection scores are optimized jointly. For $K=4$, $64.1\%$ AUC-ROC is obtained for our method while Rand-1 only performs slightly better than random (50\% AUC-ROC). In summary, the proposed AGD-based solution outperforms the traditional $l_1$ score-based solution (Rand-1) in identical white box attack settings.

\begin{figure}[t]%
\subfloat[$\alpha_a$]{
\centering
\includegraphics[width=0.5\linewidth]{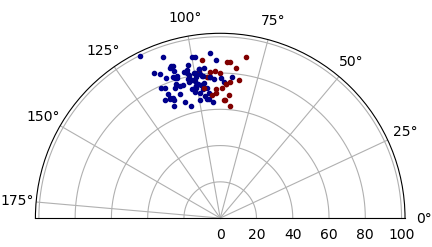}}
\centering
\subfloat[$r$]{
\includegraphics[width=0.5\linewidth]{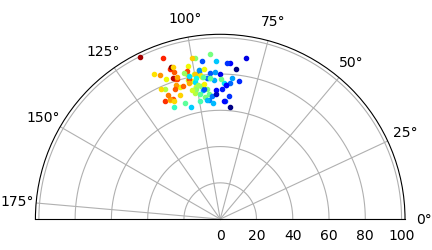}}
\caption{Variations in detection scores obtained for 100 different random transformations on an adversarial example. Scores $\alpha_a$ generated by our method are more consistent than traditional $l_1$-based scores $r$.}

\label{f110}
\end{figure}

\textbf{Detection visualization:} To investigate the consistency of the proposed score $\alpha_a$ is compared to the detection score $r$ used in Rand-1, we visualize their variations for 100 different random transformations on a random adversarial example in Fig. \ref{f110}. The color of dots correspond to min-max normalized score values. The angle is indicative of the cosine distance between vectors $f^m(\textbf{I}^{p}) -  f^m(\textbf{I}^q)$ and $\Delta f^m(\textbf{I}^p,a)$. The distance of each dot from the origin corresponds to the magnitude of vector $\Delta f^m(\textbf{I}^p,a)$. The plot shows that the proposed scores: $\alpha_a$ is more consistent than $r=||\net^m(\textbf{I}^p)  - \net^m(\textbf{I}^q)||_1$ under multiple random transformations, and the majority of the samples have low similarities. 

\section{Conclusions}
We propose the use of adversarial gradient directions for adversarial example detection. The proposed approach uses a single transformation of the input example and a single example of the predicted class from a reference database. Despite its simplicity, the proposed method has significant discriminative power and outperforms existing state-of-the-art adversarial example detection methods on CIFAR-10 and ImageNet databases in both gray-box and white-box settings. We encourage the research community to further explore the utility of adversarial gradient directions for adversarial example detection.

\section{Acknowledgements}
The authors would like to thank Dr. Yizhen Wang for his feedback on an earlier version of of this paper.

{\small
\bibliography{egbib}
}
\end{document}